# Fuzzy Classification of Facial Component Parameters

S. Halder[1], D. Bhattacherjee[2], M. Nasipuri[2], D. K. Basu[2*] and M. Kundu[2]
[1] Department of Computer Science and Engineering, RCCIIT, Kolkata -, India
Email: sant.halder@gmail.com
[2]Department of Computer Science and Engineering, Jadavpur University, Kolkata, 700032, India
Email: {debotosh@indiatimes.com, mita_nasipuri@gmail.com, dipakkbasu@gmail.com, mkundu@cse.jdvu.ac.in}
[*]AICTE Emeritus Fellow

*Abstract*—This paper presents a novel type-2 Fuzzy logic System to define the Shape of a facial component with the crisp output. This work is the part of our main research effort to design a system (called FASY) which offers a novel face construction approach based on the textual description and also extracts and analyzes the facial components from a face image by an efficient technique. The Fuzzy model, designed in this paper, takes crisp value of width and height of a facial component and produces the crisp value of Shape for different facial components. This method is designed using Matlab 6.5 and Visual Basic 6.0 and tested with the facial components extracted from 200 male and female face images of different ages from different face databases.

*Index Terms*—facial components, shape, Fuzzification, rules, inference engine, output process

## I. INTRODUCTION

The FASY system [1, 3, 5, 6] is being developed for generation of a new face from the textual description of human resources with the stored facial components extracted from different face databases. One of the main features of the FASY system is to extract the facial components from a face image and store them in the database along with their characteristics. Suppose the Shape of a facial component FC-1 can be of three types: T1, T2 and T3. Sometimes it is difficult to conclude that FC-1 belongs to only the set of T1 or T2 or T3. Rather, in this situation, this is more relevant to say that FC-1 is X% of T1 and Y% of T2 and Z% of T3. Moreover such type of characterization also gives more accurate classification of a component. To give a solution in this direction, we have designed a Fuzzy Rule Based Classifier for the analysis of the facial components after extracting them from a face image. This paper concentrates on the Fuzzy classification for the characterization of the facial components which corresponds to the most significant characteristics of the FASY system. The extraction process of our work is reported in [5, 6].

The Fuzzy model, designed in this paper, takes crisp value of width and height of extracted facial components and produces the crisp value of Shape for different facial components. These crisp values are stored into the database for later use.

A variety of face databases have already been described in literature. Here we have worked with the databases from CMU face database [2] and our own face database (DB-JU) with color (RGB) face images.

## II. FACE COMPONENTS AND DERIVED PARAMETERS

In this paper six facial components have been considered to describe a face. They are Right Eye, Right Eyebrow, Left Eye, Left Eyebrow, Nose and Lip. Each facial component has its corresponding width (*W*) and height (*H*) as shown in Fig. 1.

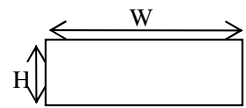

Figure 1. Concept of Width and Height of a facial component.

The present system computes the ratio H/W (labeled as *HBW*) for eyebrow, eye and lip and also calculates the value of W/H (labeled as *WBH*) for nose as shown in equation (1) and (2).

$$HBW = \frac{H}{W} \qquad (1)$$

$$WBH = \frac{1}{HBW} \qquad (2)$$

These computed values of *HBW* and *WBH* are now multiplied by 100 to find their respective percentage values using equation (3) and (4).

$$HBWP = HBW \times 100 \qquad (3)$$

$$WBHP = WBH \times 100 \qquad (4)$$

Now five values of *HBWP* are calculated for five facial components (Right Eye, Right Eyebrow, Left Eye, Left Eyebrow and Lip) and one *WBHP* value is calculated for one facial component (Nose). These crisp values of *HBWP* and *WBHP* are fed into the fuzzy system which outputs the shape description of respective facial component. The different parameters of shapes for different facial components, those we have considered in the present work, are shown in Table I.

TABLE I: TYPE OF SHAPE FOR DIFFERENT FACIAL COMPONENTS

| Components | Shape Parameters |
|---|---|
| Left Eye and Right Eye | Very Large, Large, Normal, Wide, Very Wide |
| Left Eyebrow and Right Eyebrow | Very Round, Round, Wavy, Flat, Very Flat |
| Nose | Very Narrow, Narrow, Normal, Wide, Very Wide |
| Lip | Very Linear, Linear, Low Linear, Wavy, Very Wavy |

III. FUZZY MODEL

The Fuzzy model designed in this paper is shown in Fig. 2. It takes the widths and heights of the different facial components as the input. Then it calculates the derived parameters (HBW, HBWP for eye, eyebrow, lip and WBH, WBHP for nose) for the facial components using equation (1), (2), (3) and (4). The three values of HBWP for eye, eyebrow, lip and the WBHP value for nose are fed into the Type-2 Fuzzy Logic System which produces the crisp values of Shape. Then the model finds the Degree of Membership (DOM) values for the different parameters of Shape for the each facial component and multiplies the DOM values with 100 to find the corresponding percentage value as the Shape description of the facial components.

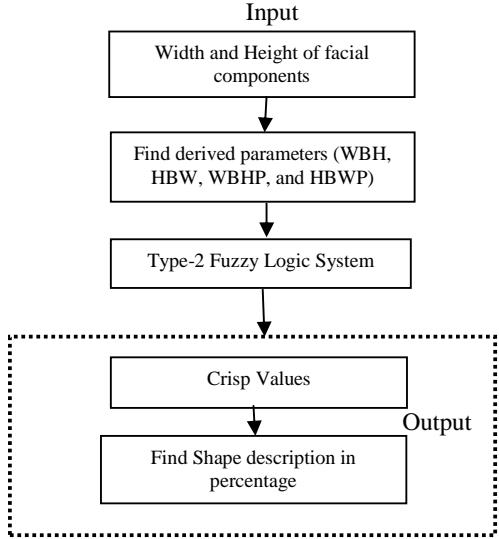

Figure 2. The Fuzzy model

IV. FUZZY LOGIC

Fuzzy Logic (FL) was first introduced by Dr. Lotfi A. Zadeh in 1962 [4]. In 1975, the motivation of type-2 fuzzy set was extended by Zadeh [7] of an ordinary type-1 fuzzy set. Type-1 fuzzy sets have certain and crisp value for their membership. While type-2 fuzzy sets do not have any certain membership and their membership is a fuzzy member. As Type-2 fuzzy system is one of the most successful techniques to deal with high uncertainty, sensitive and non-linear problem [8], hence Type-2 fuzzy system is very suitable for our research problem. The most commonly used membership functions are Triangular, Trapezoidal, and Gaussian. In the present work, we have used the Triangular membership function.

The design of novel type-2 fuzzy logic system for the classification of shape parameters for different facial components is consist of four steps, i.e. Fuzzification, Rules, Inference engine and Output process as shown in Fig. 3.

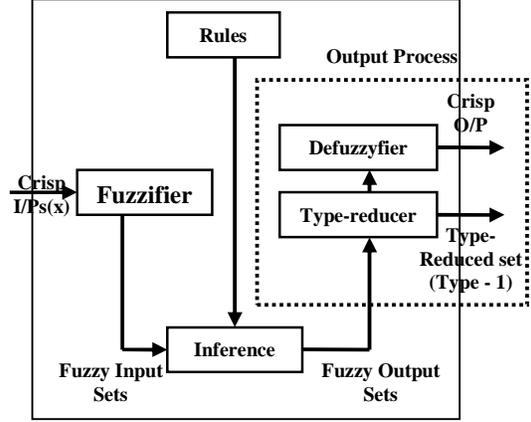

Figure 3. General model of Fuzzy Rule Based FL

*A. Fuzzification*

The process of describing crisp values into linguistic terms (fuzzy set) is called Fuzzification. Firstly the input (*HBWP*, *WBHP*) for each facial component is fuzzified into "Very Low", "Low", "Normal", "High" and "Very High" type-1 membership functions as shown in Fig. 4 to Fig. 7 for the different facial components.

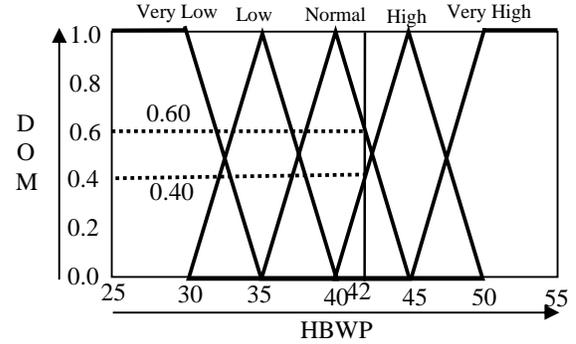

Figure 4. Type-1 fuzzy sets to represent input value HBWP for Eye

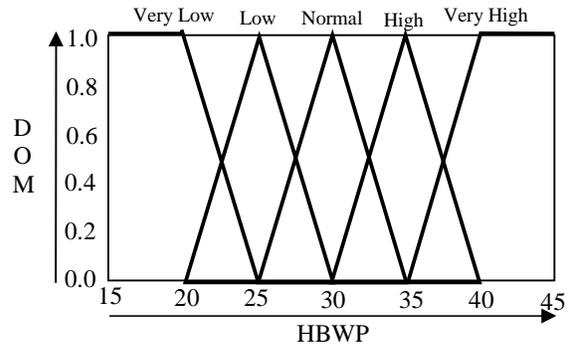

Figure 5. Type-1 fuzzy sets to represent input value HBWP for Eyebrow

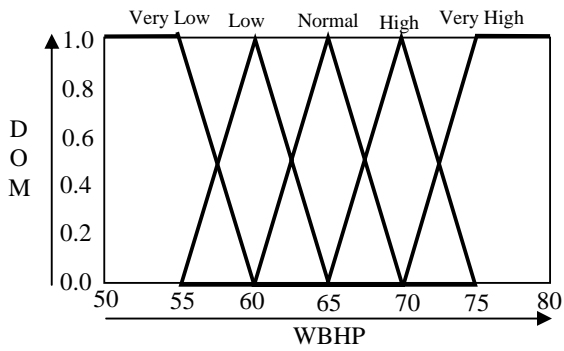

Figure 6. Type-1 fuzzy sets to represent input value WBHP for Nose

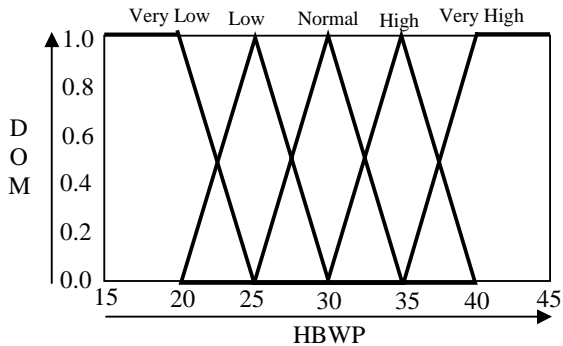

Figure 7. Type-1 fuzzy sets to represent input value HBWP for Lip

Secondly the output Shapes of each facial component are represented into type-2 membership functions as shown in Fig. 8 to Fig. 11 for the different facial components.

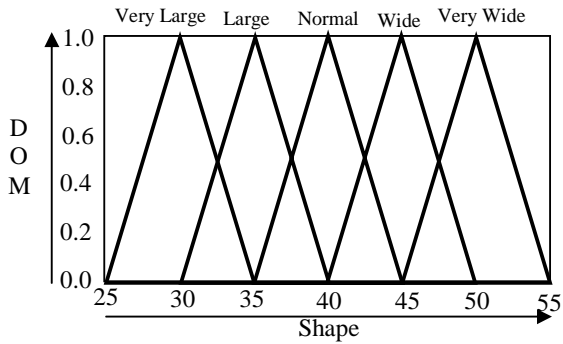

Figure 8. Type-2 fuzzy sets to represent output value Shape for Eye

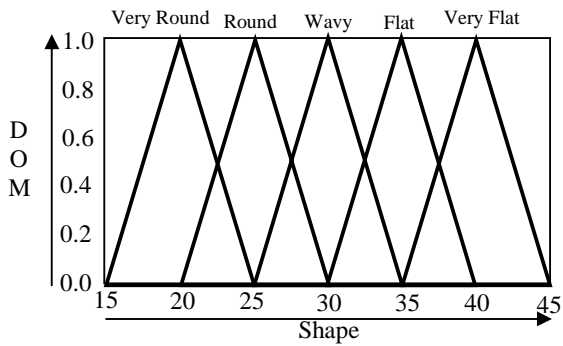

Figure 9. Type-2 fuzzy sets to represent output value Shape for Eyebrow

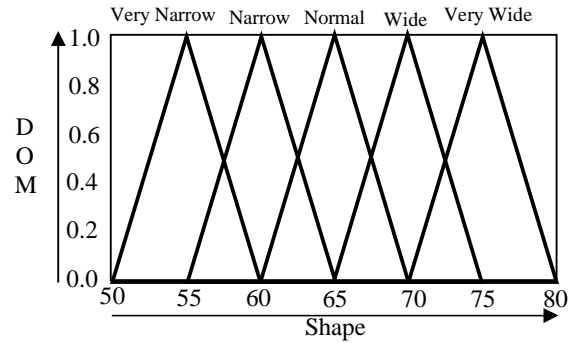

Figure 10. Type-2 fuzzy sets to represent output value Shape for Nose

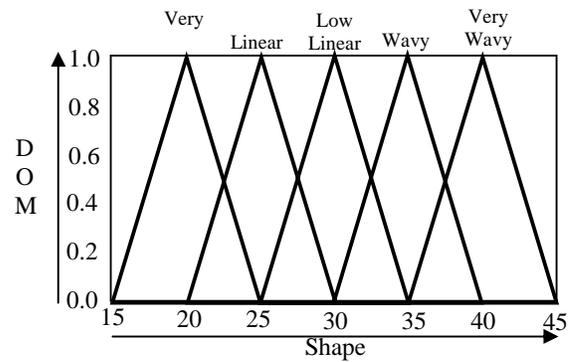

Figure 11. Type-2 fuzzy sets to represent output value Shape for Lip

*B. Rules*

FLS rules are extracted from the data that depend on the conducted experiments to map input with output. Each rule has two parts; "IF" part which is called antecedent and "THEN" part which is called consequent.

The extracted FLS rules for eye are given below:
Rule1: IF HBWP is "Very Low" Then SHAPE is "Very Large".
Rule2: IF HBWP is "Low" Then SHAPE is "Large".
Rule3: IF HBWP is "Normal" Then SHAPE is "Normal".
Rule4: IF HBWP is "High" Then SHAPE is "Wide".
Rule5: IF HBWP is "Very High" Then SHAPE is "Very Wide".

The extracted FLS rules for eyebrow are given below:
Rule1: IF HBWP is "Very Low" Then SHAPE is "Very Round".
Rule2: IF HBWP is "Low" Then SHAPE is "Round".
Rule3: IF HBWP is "Normal" Then SHAPE is "Wavy".
Rule4: IF HBWP is "High" Then SHAPE is "Flat".
Rule5: IF HBWP is "Very High" Then SHAPE is "Very Flat".

The extracted FLS rules for nose are given below:
Rule1: IF WBHP is "Very Low" Then SHAPE is "Very Narrow".
Rule2: IF WBHP is "Low" Then SHAPE is "Narrow".
Rule3: IF WBHP is "Normal" Then SHAPE is "Normal".
Rule4: IF WBHP is "High" Then SHAPE is "Wide".

Rule5: IF WBHP is "Very High" Then SHAPE is "Very Wide".

The extracted FLS rules for lip are given below:
Rule1: IF HBWP is "Very Low" Then SHAPE is "Very Linear".
Rule2: IF HBWP is "Low" Then SHAPE is "Linear".
Rule3: IF HBWP is "Normal" Then SHAPE is "Low Linear".
Rule4: IF HBWP is "High" Then SHAPE is "Wavy".
Rule5: IF HBWP is "Very High" Then SHAPE is "Very Wavy".

*C. Inference Engine*

The fuzzy inference engine applies the fuzzy rules on truth value of input variables in order to determine the corresponding output. The most commonly used methods are minimum t-norm and product t-norm. In this paper, the product t-norm inference implication method is used to scale the output membership function Shape by the truth values of HBWP/WBHP for each facial components. Now suppose the HBWP value of an Eye labeled as Eye-1 is 42. For HBWP = 42, the type-1 fuzzy set for Eye-1 has non-zero membership value in two membership functions, i.e. 0.40 of "High" and 0.60 of "Normal" as shown in Fig. 4. Now the implementation of fuzzy rules and product t-norm method generate two corresponding membership values of Shape membership functions for Eye as shown in Fig. 12.

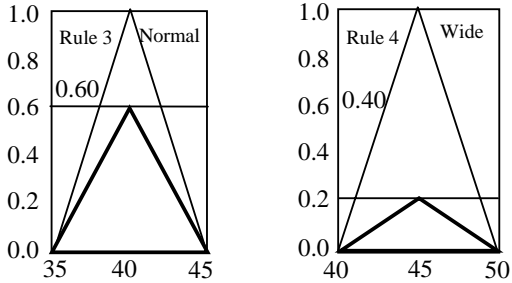

Figure 12. Implementation of fuzzy rules

Now the product t-norm method is implemented on two corresponding membership values of Shape of Eye as shown in Fig. 13.

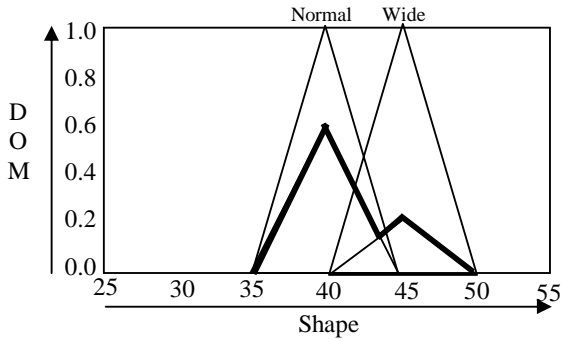

Figure 13. Implementation of product t-norm method

*D. Output Process*

Firstly the defuzzifier uses the centroid method to find the crisp value of Shape. For Eye-1, our Fuzzy model produces the crisp value 42.097 as shown in Fig. 14.

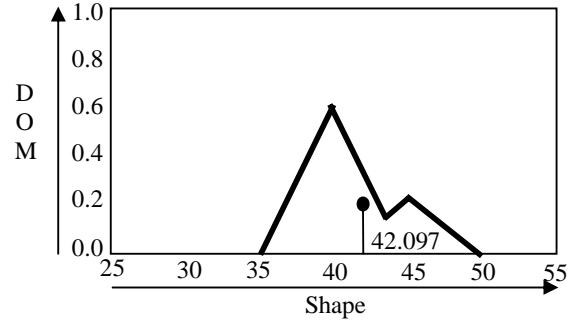

Figure 14. Implementation of "centroid" defuzzification method

Secondly based on the crisp value, the system calculates the degree of membership values for all the parameters of each Shape for each facial component and multiplies the DOM values by 100 to find the corresponding percentage value. For Eye-1, the values of the different parameters of Shape are shown in Fig. 15 and these percentages values are stored in the database as the Shape description of the Eye-1 for later use.

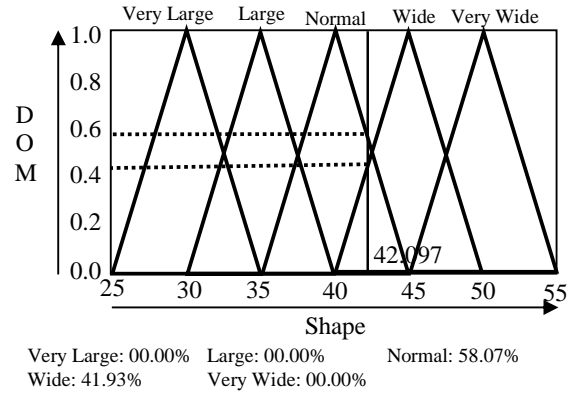

Very Large: 00.00%   Large: 00.00%   Normal: 58.07%
Wide: 41.93%         Very Wide: 00.00%

Figure 15. Final output for Shape description in percentage

V. EXPERIMENTAL RESULTS

For testing the proposed method for Fuzzification of the shape of the facial components, we had about 200 male and female face images of different ages. The images are collected from the CMU database and DB-JU database. Fig. 16 shows some face images collected from different face databases. Fig. 17 to Fig. 20 show the facial components extracted from the faces of Fig. 16. The calculated values of the different parameters of Shape for facial components, shown in Fig. 17 to Fig. 20, are illustrated in Table II. Table II depicts the percentage values of those parameters of Shape for each facial component that have non zero values.

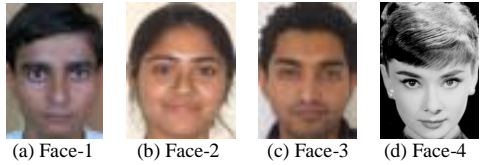
(a) Face-1  (b) Face-2  (c) Face-3  (d) Face-4

Figure 16. (a), (b) and (c) Face images from DB-JU Database (d) Face image from CMU Database

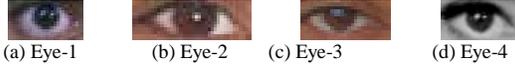
(a) Eye-1  (b) Eye-2  (c) Eye-3  (d) Eye-4

Figure 17. Extracted right eyes from the face images given in Fig. 16

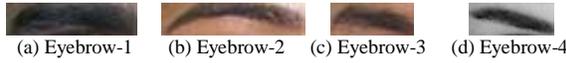
(a) Eyebrow-1  (b) Eyebrow-2  (c) Eyebrow-3  (d) Eyebrow-4

Figure 18. Extracted right eyebrows from the face images of Fig. 16

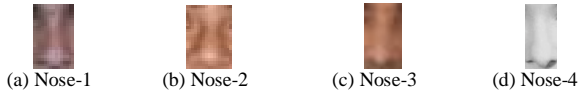
(a) Nose-1  (b) Nose-2  (c) Nose-3  (d) Nose-4

Figure 19. Extracted noses from the face images of Fig. 16

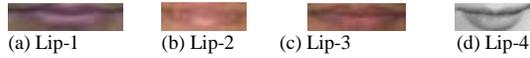
(a) Lip-1  (b) Lip-2  (c) Lip-3  (d) Lip-4

Figure 20. Extracted lips from the face images of Fig. 16

TABLE II: TYPE OF SHAPE FOR DIFFERENT FACIAL COMPONENTS

| Facial Components | W | H | HBWP/WBHP | Shape |
|---|---|---|---|---|
| Eye-1 | 48 | 24 | 50.00 | Wide : 00.05 %<br>Very Wide : 99.95% |
| Eye-2 | 38 | 12 | 31.57 | Very Large : 65.31%<br>Large : 34.69% |
| Eye-3 | 27 | 10 | 37.03 | Large : 57.55%<br>Normal : 42.45% |
| Eye-4 | 39 | 20 | 51.28 | Wide : 00.06%<br>Very Wide : 99.94% |
| Eyebrow-1 | 63 | 17 | 26.98 | Round : 58.37%<br>Wavy : 41.63% |
| Eyebrow-2 | 63 | 14 | 22.22 | Very Round : 54.50%<br>Round : 45.50% |
| Eyebrow-3 | 63 | 14 | 38.88 | Flat : 26.47%<br>Very Flat : 73.53% |
| Eyebrow-4 | 49 | 19 | 38.77 | Flat : 28.59%<br>Very Flat : 71.41% |
| Nose-1 | 49 | 84 | 58.33 | Very Narrow : 36.40%<br>Narrow : 63.60% |
| Nose-2 | 47 | 69 | 68.11 | Normal : 40.10%<br>Wide : 59.90% |
| Nose-3 | 33 | 59 | 55.93 | Very Narrow : 77.26%<br>Narrow : 22.76% |
| Nose-4 | 35 | 68 | 51.47 | Very Narrow : 99.96%<br>Narrow : 00.04% |
| Lip-1 | 65 | 17 | 26.15 | Linear : 72.92%<br>Low Linear : 27.08% |
| Lip-2 | 63 | 20 | 31.74 | Low Linear : 62.41%<br>Wavy : 37.59% |
| Lip-3 | 46 | 13 | 28.26 | Linear : 37.60%<br>Low Linear : 62.40% |
| Lip-4 | 60 | 21 | 35.00 | Wavy : 99.96%<br>Very Wavy : 00.04% |

ACKNOWLEDGMENT

Authors are thankful to the "Center for Microprocessor Application for Training Education and Research", "Project on Storage Retrieval and Understanding of Video for Multimedia" of Computer Science & Engineering Department, Jadavpur University, for providing infrastructural facilities during progress of the work. One of the authors, Mr. Santanu Halder, is thankful to RCC Institute of Information Technology for kindly permitting him to carry on the research work and Dr. D. K. Basu acknowledges the thanks to AICTE, New Delhi for providing an Emeritus fellowship.